%% file: main.tex
\documentclass[10pt,journal,compsoc]{IEEEtran}
\usepackage{amsmath,amsfonts}
\usepackage{amsthm,amssymb}
\usepackage{amsmath,amssymb,amsfonts,dsfont,pifont,bm,bbm,mathrsfs,mathtools,nicefrac}
\usepackage{xcolor,algorithm,listings}
\usepackage{algorithmicx}
\usepackage{algpseudocode}
\usepackage{array}
\usepackage[caption=false,font=normalsize,labelfont=sf,textfont=sf]{subfig}
\usepackage{textcomp}
\usepackage{booktabs,multirow,adjustbox,diagbox,threeparttable,multicol}
\usepackage{colortbl}
\usepackage{stfloats}
\usepackage{url}
\usepackage{multirow}
\usepackage{makecell}
\usepackage{caption}
\usepackage{verbatim}
\usepackage{graphicx}
\usepackage{bbding}
\usepackage{xspace}
\usepackage{cite}
\usepackage{xcolor}
\def\BibTeX{{\rm B\kern-.05em{\sc i\kern-.025em b}\kern-.08em
    T\kern-.1667em\lower.7ex\hbox{E}\kern-.125emX}}

\usepackage[pagebackref,breaklinks,colorlinks,linkcolor=blue,citecolor=blue, bookmarks=false]{hyperref}
\usepackage[capitalize]{cleveref}  
\Crefname{section}{Section}{Sections}
\Crefname{table}{Table}{Tables}
\Crefname{figure}{Figure}{Figures}
\Crefname{equation}{Equation}{Equations}
\definecolor{codegreen}{rgb}{0,0.6,0}
\definecolor{codegray}{rgb}{0.5,0.5,0.5}
\definecolor{codepurple}{rgb}{0.58,0,0.82}
\definecolor{backcolour}{rgb}{1.0,1.0,1.0}
\lstdefinestyle{mystyle}{
    backgroundcolor=\color{backcolour},
    commentstyle=\color{codegreen},
    keywordstyle=\color{magenta},
    numberstyle=\tiny\color{codegray},
    stringstyle=\color{codepurple},
    basicstyle=\ttfamily\scriptsize,
    breakatwhitespace=false,
    breaklines=true,
    captionpos=b,
    keepspaces=true,
    numbers=left,
    numbersep=5pt,
    showspaces=false,
    showstringspaces=false,
    showtabs=false,
    tabsize=2
}
\definecolor{one_shot}{RGB}{220, 227, 243}
\definecolor{few_shot}{RGB}{230, 242, 219}
\newcommand{\hut}{\textcolor{black}}
\newcommand{\yr}{\textcolor{black}}

\newcommand{\yyy}[1]{{\textcolor{black}{#1}}}

\begin{document}

\title{MotionMaster: Training-free Camera Motion Transfer For Video Generation}

\author{Teng Hu$^\ast$, Jiangning Zhang$^\ast$, Ran Yi$^\dag$, Yating Wang, Hongrui Huang,\\ Jieyu Weng, Yabiao Wang, Lizhuang Ma
\IEEEcompsocitemizethanks{
\IEEEcompsocthanksitem Teng Hu, Ran Yi, Yating Wang, Jieyu Weng, and Lizhuang Ma are with the Department of Computer Science and Engineering, Shanghai Jiao Tong University, Shanghai 200240, China (email: \{hu-teng,ranyi,wyating\_0929,w.jerry,ma-lz\}@sjtu.edu.cn).
\IEEEcompsocthanksitem Jiangning Zhang and Yabioa Wang are with the Youtu Lab, Tencent, Shanghai 200233, China (email: \{vtzhang, caseywang\}@tencent.com).
\IEEEcompsocthanksitem Hongruihuang is with the Faculty of Computing, Harbin Institute of Technology, Harbin 150001,  (email: 2022212016@stu.hit.edu.cn).
}
}



\IEEEtitleabstractindextext{
\begin{abstract}
The emergence of diffusion models has greatly propelled the progress in image and video generation. 
Recently, 
\yr{some efforts have been made in controllable video generation, including text-to-video, image-to-video generation, video editing, and video motion control, among which camera motion control is an important topic.}
%
%
However, 
existing \yr{camera motion control} methods rely on training 
\yr{a temporal camera module, and}
necessitate substantial computation resources
due to the large amount of parameters in video \yr{generation} models.
Moreover, \yr{existing methods} pre-define camera motion \yr{types} during training\yr{, which} limits their flexibility in camera control, preventing the realization of some specific camera controls, such as various camera movements in films. 
Therefore, to reduce training \yr{costs} and achieve flexible camera control, we propose \yr{MotionMaster,} a novel training-free video motion transfer model, 
which 
\yr{disentangles camera motions and object motions in source videos,} 
and \yr{transfers} the extracted camera motion\yr{s to} new video\yr{s}. 
%
We first \yr{propose} a \yr{one-shot camera motion disentanglement}
method \yr{to extract camera motion from} a single source video, \yr{which} separate\yr{s the} moving objects from the background and estimate\yr{s the} camera motion in \yr{the moving objects region based on the motion in the background} by solving a Poisson equation. 
Furthermore, we propose a \yr{few-shot camera motion disentanglement method \yr{to} extract \yr{the} common camera motion \yr{from multiple videos with similar camera motions}, which employs a} window-based clustering technique \yr{to} extract \yr{the} common 
\yr{features in temporal attention maps of multiple videos.}
%
\hut{Finally, we \yr{propose} a motion combination method to combine different types of camera motions together, enabling our model a more controllable and flexible camera control.} 
Extensive experiments demonstrate that our training-free approach can effectively decouple camera-object motion and apply the decoupled \yr{camera} motion to a wide range of controllable video generation tasks, achieving flexible and diverse camera motion control. More details can be referred in
\url{https://sjtuplayer.github.io/projects/MotionMaster}.
\end{abstract}
}
\maketitle
\begin{figure*}[t]
  \includegraphics[width=\textwidth]{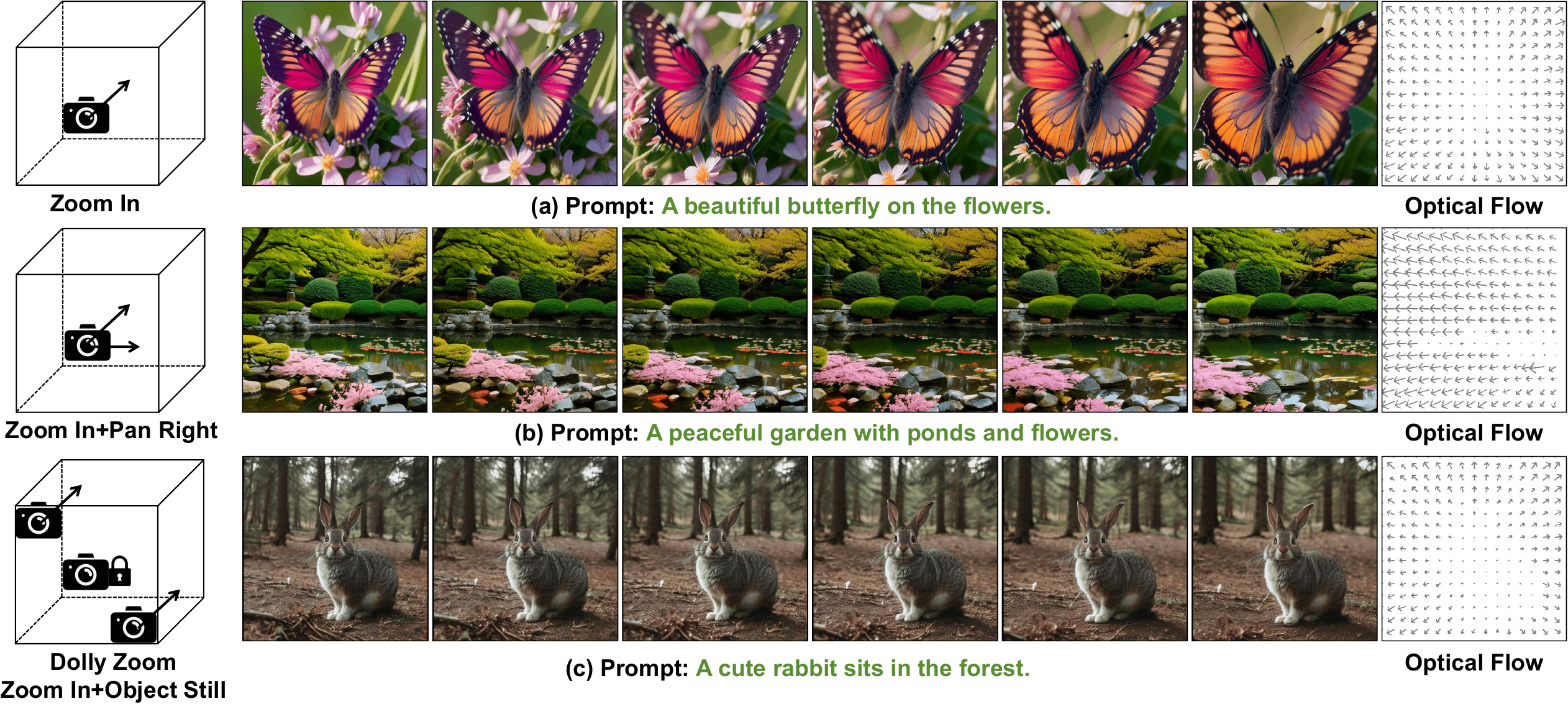}
  \caption{Flexible and diverse camera motion control of our training-free MotionMaster. MotionMaster can control (a) one camera motion or (b) combine several camera motions in one video. Moreover, MotionMaster enables control of different camera motions in different regions, which can achieve professional Dolly Zoom with zooming motions in the background and fixed motion in the foreground (c).
  }
  \label{fig:teaser}
\end{figure*}

\input{secs/1_introduction}

\input{secs/2_related_work}

\input{secs/3_method}

\input{secs/4_experiment}

\input{secs/5_conclusion}

 {
    \small
    \bibliographystyle{IEEEtran}
    \bibliography{main}
}

\clearpage
\appendix
\input{secs/6_suppl}

\end{document}

%% file: secs/1_introduction.tex
\section{Introduction}

In recent years, the rapid development of generative models~\cite{ho2020ddpm,rombach2022ldm} has led to significant advancements in the field of image and video generation. 
Among video generation, diffusion models~\cite{guo2023animatediff,blattmann2023svd,singer2022make-a-video,chen2024videocrafter} have emerged as powerful tool\yr{s} for generating high-quality videos with high diversity. 
Meanwhile, the demand for controllable video generation has grown \yr{significantly}, 
especially in \yr{applications} such as film production, virtual reality, and video games, \hut{where researchers have devoted \yr{much} effort to \yr{controllable generation} tasks \yr{including} 
text-to-video \yr{generation}~\cite{guo2023animatediff,chen2024videocrafter,chen2023videocrafter1,singer2022make-a-video}, 
image-to-video \yr{generation}~\cite{blattmann2023svd,guo2023animatediff}, 
video motion control~\cite{chen2023motion,tu2023motioneditor,chen2024motion_zero,yang2024direct_a_video}, 
and video editing~\cite{bai2024uniedit,qi2023fatezero}. 
Since video is composed of a sequence of images with consistent and fluent motions, \yr{the control of} video motion has become 
\yr{an important topic in} controllable video generation.}

For \hut{video} motion control, \textbf{1)} most of the existing methods~\cite{chen2024motion_zero,chen2023motion,wu2023tune,bai2024uniedit} focus on modeling the \textit{object motion} and use trajectory or a source video to guide the movement of the objects, but usually lack the ability to model the camera motion. 
\textbf{2)} To enable the control of the \textit{camera motion}, AnimateDiff~\cite{guo2023animatediff} trains temporal LoRA modules~\cite{hu2021lora} on a collected set of videos with the same camera motion. 
To control different camera motions \yr{using} one model, MotionCtrl~\cite{wang2023motionctrl} labels a large number of videos with corresponding camera pose parameters to train a camera motion control module.
In contrast, Direct-a-video~\cite{yang2024direct_a_video} utilizes a self-supervised training process by manually constructing camera motions along x, y, and z axis, reducing the training resources to some extent. 
However, all the existing \yr{camera motion control} methods rely on training a temporal camera module to control the camera motion, which poses a significant requirement to the computational resources 
\yr{due to the large number of parameters in video generation models.}
Moreover, these methods can only achieve simple camera motion control and cannot handle some complex \yr{and} professional camera motions in films, \yr{such as} Dolly Zoom (zoom in or out the camera while keeping the object still) and Variable-Speed Zoom (zoom with variable speed).

To achieve complex camera motion control and reduce the training \yr{costs}, we propose MotionMaster, a novel \textbf{training-free camera motion \yr{transfer} model},
which 
\yr{disentangles camera motions and object motions in} source videos 
and then \yr{transfers} the extracted \yr{camera} motion\yr{s} to new videos. First\yr{ly}, we 
\yr{\yyy{observe} that the temporal attention maps in diffusion-based video generation models contain the information of video motions, and find that the motions are composed of two motion types, camera motions and object motions.}
\hut{\yr{We then} propose two methods to disentangle the camera motions and object motions in temporal attention maps.}
\yr{1) In} \textbf{one-shot camera motion \yr{disentanglement}}, \yr{we decompose camera and object motions from a single source video.} 
We regard the motion in \yr{the} background as only containing camera motion, while the motion in the foreground as containing both camera and object motions.
\yr{W}e employ a segmentation model to 
\yr{separate} the moving objects and background regions,
and then predict the camera motion in foreground region from background motion by solving a Poisson equation. 
\yr{2)} To further enhance the disentanglement ability, we propose a \textbf{few-shot camera motion \yr{disentanglement}} method \yr{to extract the common camera motion from several videos with similar camera motions}, which employs a novel window-based clustering method to extract \yr{the} common 
\yr{features from temporal attention maps of multiple videos.}
\hut{Finally, we investigate the additivity and positional composition ability of camera motion\yr{s}, and propose a camera motion combination method to achieve flexible camera control, which can enable \yr{combining} different kinds of camera motions into a new motion\yr{, and apply} different camera motions in different \yr{regions}.}

Extensive experiments demonstrate the superior \yr{performance} of our model in both one-shot and few-shot camera motion transfer. 
With the camera motion \yr{combination} and the disentanglement between the camera motion and position, our model 
\yr{substantially improve the controllability and flexibility of camera motions.}

The main contributions can be summarized as follows:

\begin{itemize}
    \item We propose MotionMaster, a training-free camera motion transfer method based on Camera-Object Motion Disentanglement, which can transfer the camera motion from source videos to newly generated videos.
    \item We propose a novel one-shot camera-object motion disentanglement method. B\hut{y \yr{separating} the moving object\yr{s} and the background \yr{regions} and estimating the camera motion in the moving object\yr{s region} by solving a Poisson equation}, our model can effectively disentangle the camera motion from object motion in a single video.
    \item 
    We \yr{further} propose a few-shot camera-object motion disentanglement method, \hut{which employs a novel window-based clustering method to extract the common camera motion from several given videos \yr{with similar camera motions}}, 
    \yr{effectively dealing with scenarios with overly complex and diverse object motions.}
    \item \hut{We \yr{propose} a camera motion combination method to achieve flexible camera motion control, which enables the model to combine different camera motions into a new motion and apply different camera motions in different \yr{regions}.}  
\end{itemize}

%% file: secs/2_related_work.tex
\section{Related Work}
\subsection{Text-to-Video Generation}

Generative models have rapidly advanced and achieved tremendous success in text-driven video generation tasks, which mostly rely on generative adversarial networks (GANs)~\cite{vondrick2016generating, wang2019few, saito2017temporal, zhang2020dtvnet} and diffusion models~\cite{girdhar2023emu, blattmann2023align, chen2023videocrafter1, ho2022video, guo2023animatediff, blattmann2023svd, singer2022make-a-video, chen2024videocrafter}
Among these methods, diffusion models have emerged as a powerful tool due to their ability to generate diverse and high-quality contents. 
Early text-driven video generation models \cite{ho2022imagen, ho2022video, singer2022make-a-video} perform diffusion in pixel space, requiring cascaded generation and significant computational resources to generate high-resolution videos. Recent research papers have implemented diffusion in the latent space \cite{rombach2022high, blattmann2023align, guo2023animatediff, zhou2022magicvideo, wang2023modelscope, blattmann2023svd}, achieving high-quality and long-duration video generation.
Additionally, researchers are exploring more controllable video generation approaches. For instance, \cite{chen2023control, esser2023structure, guo2023sparsectrl} introduce spatial and geometric constraints to generative models, \cite{wei2023dreamvideo} generates videos of desired subject, and \cite{wang2023motionctrl, chen2023motion} govern motion in generated videos. These methods enable users to finely control various attributes of videos, resulting in generated outcomes that better align with user preferences and requirements.

%
\subsection{Motion Controllable Video Generation}
\noindent\textbf{Object Motion Control.} 
Many researches~\cite{jeong2023vmc,jain2023peekaboo, teng2023drag, tu2023motioneditor, wu2023lamp, zhao2023motiondirector, wei2023dreamvideo} have been conducted to control object motions to better align with user preferences.
Some methods~\cite{chen2024motion_zero,yang2024direct_a_video} enable users to control the motion of objects by dragging bounding boxes, while some other works \cite{wang2023motionctrl,jain2023peekaboo} allow control over the trajectory of the object. VideoComposer~\cite{wang2024videocomposer} provides global motion guidance by conditioning on pixel-wise motion vectors.
Besides, some video editing methods \cite{deng2023dragvideo, tu2023motioneditor, bai2024uniedit, qi2023fatezero} also enable motion editing through text-driven or manually specified motions, which requires motion consistency between adjacent frames. 
In summary, all these works focus more on controlling the object motions rather than camera motions, which operates at a local, high semantic level.

\noindent\textbf{Camera Motion Control.} 
\yyy{There have been relatively few researches in camera motion control.}
AnimateDiff~\cite{guo2023animatediff} employs temporal LoRA modules~\cite{hu2021lora} trained on a collected set of videos with similar camera motion. Thus a single LoRA module is capable of controlling only a specific type of camera motion. 
MotionCtrl~\cite{wang2023motionctrl} constructs a video dataset annotated with camera poses to learn camera motions, but requires substantial manual effort.
Direct-a-video~\cite{yang2024direct_a_video} adds camera motion along coordinate axes to existing videos, which can reduce annotation costs. However, all of these works require fine-tuning pretrained video generation models, consuming a large amount of computation resources and limiting the style of camera motion to the training data. In contrast, our model enables flexible camera motion control with any target camera motions without re-training the model, which brings a much wider application for camera control in video generation. 



%% file: secs/3_method.tex
\begin{figure*}[t]
\centering
\includegraphics[width=0.85\textwidth]{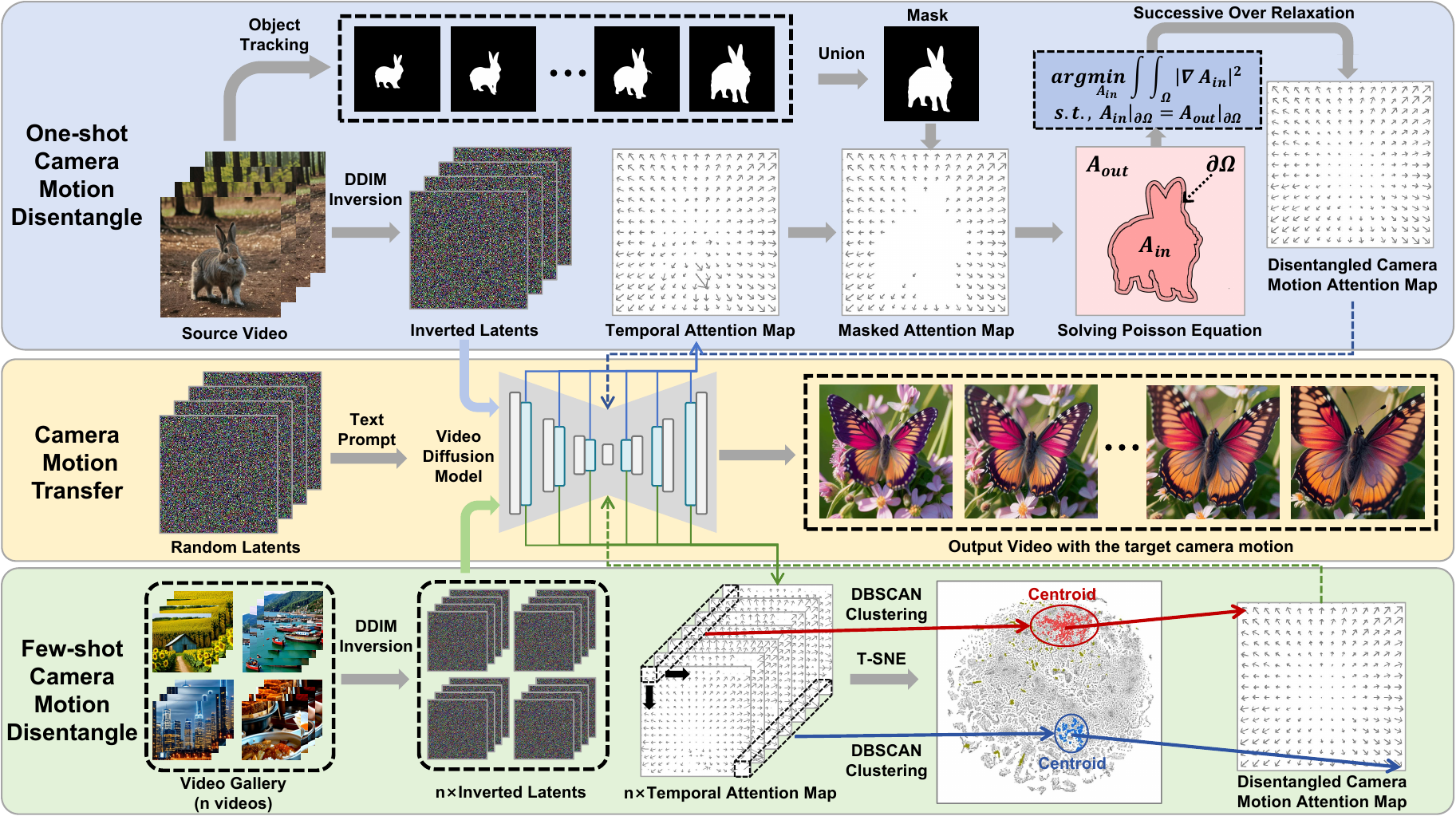}
\vspace{-0.1in}
\caption{\textit{Main framework of our method:} Our model can extract the camera motion from a single video or several videos that share similar camera motions. 
\colorbox{one_shot}{1) \textit{One-shot camera motion disentanglement:}} We first employ \hut{SAM~\cite{kirillov2023segment_any_thing}} to \yyy{segment} the moving objects in the source video and extract the temporal attention maps from the inverted latents. To disentangle the camera and object motions, we mask out the object areas of the attention map and estimate the camera motion inside the mask by solving a Poisson equation\yyy{.}
\colorbox{few_shot}{2) \textit{Few-shot camera motion disentanglement:}} we extract \yyy{the common camera motion from} the temporal attention maps \yyy{of} several given videos. For each position $\mathbf{(x,y)}$, we employ all of its k-neighboring attention map values across each video for clustering.
Then, we use the centroid of the largest cluster to represent the camera motions in position $\mathbf{(x,y)}$.}
\label{fig:main framework}
\vspace{-0.1in}
\end{figure*}

\section{Method}

Our \yr{MotionMaster} model aims to \textbf{disentangle the camera motion and object motion} in a single or several videos, and then \textbf{transfer the disentangled camera motion} to the newly generated videos. 
We 
\yr{first} 
\yyy{observe} that 
\yr{the temporal attention maps \yyy{in diffusion-based video generation models} contain the information of videos motions\yyy{, and find that the motion are composed of two motion types, camera motions and object motions}.}
We \yr{then} propose two methods to decompose the temporal attention map $Attn$ into object motion $Attn^o$ and camera motion $Attn^c$, \hut{as shown in Fig.~\ref{fig:main framework}}. By substituting the temporal attention map with the 
\yr{temporal attention of}
the target camera motion, we can enable the video \yr{generation} models to generate videos with the desired camera motion. 

Specifically, to disentangle the camera motion from the object motion, we propose to extract the camera motions from either a single video or a few (5-10) videos. 
\textbf{1)} In \textbf{one-shot camera motion disentanglement}, \yr{we aim to extract camera motion from a single video (Fig.~\ref{fig:main framework} top).}
Considering \yr{the} motion in background region only contains camera motion, while motion in \yr{the} foreground region contains both camera motion and object motion, we first separate background and foreground regions.
We employ 
SAM~\cite{kirillov2023segment_any_thing} to segment the moving objects, and decompose the \yr{given} video into moving object \yr{region} $M$ and background \yr{region} $\tilde{M}=1-M$. 
\yr{Then} we regard the motion in \yr{the background region} $\tilde{M}$ as \yr{only containing} camera motion. 
With the observation that the camera motion is smooth and continuous, and the neighboring pixels share similar motions\hut{~\cite{zhou2023mvflow,fleet2006optical,zhang2023modeling,gharibi2023multi}}, we construct a Poisson equation to estimate the camera motions in \yr{the moving objects region} $M$ based on the given camera motions in \yr{the background region} $\tilde{M}$, achieving camera-object motion disentanglement for a single video. 

\textbf{2)} When the object motions 
are too complex to disentangle \yr{from a single video}, we \yr{propose} a \textbf{few-shot camera motion disentanglement} method to extract common camera motion from $m$ \yr{(5-10)} videos with similar camera motions \yr{(Fig.~\ref{fig:main framework} bottom)}.
\yr{To extract common camera motion of $m$ videos,} we regard the common feature of the temporal attention maps of these videos as the feature of the common camera motion.
We \yr{then} propose a window-based 
clustering method for each pixel of the temporal attention map to extract the common camera motion \yr{and filter out outliers}.
Specifically, we regard the neighboring pixels in a $k\times k$ window share similar camera motions and cluster the $k^2$-neighboring pixels of each pixel in the $m$ temporal attention maps with DBSCAN clustering method, where the centroid of \hut{the largest cluster} can be used to represent the \yr{common} camera motion. 

\hut{Finally, we investigate the additivity and positional composition ability of camera motion\yr{s}. 
We propose a camera motion combination method to achieve flexible camera motion control, which can combine different camera motions into a new motion and apply different camera motions in different \yr{regions}, 
\yr{substantially} improving the controllability and flexibility of camera motion\yr{s}.}


\subsection{Camera Motion Extraction Based on Temporal Attention}
\label{sec:cmt}

\textbf{Preliminar\yr{ies} of temporal attention module.} 
Most of the current video \yr{generation} models~\cite{blattmann2023align,blattmann2023svd,guo2023animatediff} are built on a pretrained text-to-image diffusion model~\cite{rombach2022ldm}, which employs spatial attention module to model the image generation process. 
To extend the image \yr{generation} models to generate videos, temporal attention module~\cite{blattmann2023align,guo2023animatediff} is proposed to enable the pretrained image \yr{generation} models with the ability to model the temporal relationship between each frame of the video. 
Specifically, the temporal attention mechanism is a self-attention module, which takes the feature map $f_{in}$ of $t$ frames \yr{(}$b\times t\times c\times h\times w$\yr{)}
as input, and reshapes it to a $(b\times h\times w)\times t\times c$ 
feature map $f$. 
Then, a self-attention module \yr{is employed} to \yr{capture} the temporal 
\yr{relationships between $t$ frames}, and output a feature map with temporal relationships between each frame, which is formulated as follows:
\begin{equation}
\begin{aligned}
    Attn&= Softmax(\frac{Q K^T}{\sqrt{\yyy{c}}}), ~~~
    ~~f_{out} = Attn V,
\end{aligned}
\end{equation}
where $Q=W_Q f$, $K=W_K f$ and $V=W_V f$, and $W_Q$, $W_K$ and $W_V$ are learnable query, key and value matrices. 

\textbf{Extracting motion information from temporal attention map.}
\hut{UniEdit}~\cite{bai2024uniedit} found that the temporal attention modules model the inter-frame dependency \yyy{and motion information\footnote{\yyy{Our experiments also validate this finding, shown in \#Suppl.}},} and use the temporal attention for video \yr{motion} editing tasks, 
\yr{where the global motion of video is edited guided by text.}
However, \yr{it} lack\yr{s} a deep analysis of how the temporal attention module models the \yr{inter-}frame dependency. 
In this paper, we find that the attention maps $Attn$ of the temporal attention layer are composed of \textbf{two motion types}, which are \textbf{camera motions} and \textbf{object motions}. 
We propose two methods to decouple motion in temporal attention map into camera and object motions (Sec.~\ref{sec:one_cme} and ~\ref{sec:few_cme}), where we disentangle the temporal attention map $Attn$ extracted from a video into camera \yr{motion} attention $Attn^c$ and object \yr{motion} attention $Attn^o$.
After decoupling camera motion from object motion, we can easily transfer camera motion \yr{from a source video $v_s$ to a target video $v_t$,} by \yr{replacing the temporal attention map of $v_t$ \hut{with} the temporal attention map $Attn^c_s$ \hut{that corresponds} to the camera motion of $v_s$}:
\begin{equation}
\begin{aligned}
    f_{out}=Attn^c_s V. 
\end{aligned}
\end{equation}


\subsection{One-shot Camera Motion Disentanglement}
\label{sec:one_cme}

\yr{In this section, we propose a method to disentangle camera motions from object motions in a single source video.}
\hut{\yr{A} video can usually be divided into two parts: the 
\yr{foreground region} and the background \yr{region}.}
\yr{Considering the motion in background region mainly contains camera motion, while the motion in foreground region contains both camera motion and object motion, 
we first extract camera motion in background region, and then predict camera motion in foreground based on the background camera motion.
}

Specifically, we first employ \hut{segment-anything model} to segment the moving objects \yr{and the background}, and then take the temporal attention map from the 
\yr{background region} as the camera motions. 
Based on the 
\yr{observation} that the camera motions are continuous and the neighboring pixels have similar camera motions, we construct a Poisson equation to estimate the camera motions inside the moving object \yr{region} based on the camera motions outside, \yr{thereby} achieving the camera-object motion disentanglement.

\textbf{Obtaining temporal attention map of a video by DDIM inversion.}
First of all, to obtain the temporal attention map of the source video $v_s$, we apply DDIM inversion on the source video to invert it into a $T$-step latent $x_T$. 
Then, by denoising $x_T$ with the video diffusion model, we \yr{obtain} a series of attention maps $\{Attn_T,Attn_{t-1}\cdots Attn_{1}\}$ in different timesteps. 
Different from the \yr{spatial attention maps (in} spatial attention modules\yr{), which model different spatial relationships in different timesteps,} 
the temporal attention maps model \textit{the temporal motion} of the video, and we find they are similar in different timesteps. 
Therefore, we can \yr{use} one representative temporal attention map $Attn=Attn_t$ \yr{at timestep $t$} to model the temporal motion, 
which can effectively reduce the computation resources to $\frac{1}{T}$ of 
\yr{using all timesteps.}
\hut{We \yr{adopt} a \yr{medium} timestep $t$, since when $t$ is large, there are too many noises in the video feature;
while when $t$ is small, \yr{the denoising has almost been completed} and the \yr{overall} motion \yr{has} already \yr{been} determined, \yr{thus} the motion information in the temporal attention map \yr{at small $t$} is not sufficient.}


\textbf{Extracting \yr{camera} motion in background region.}
With the obtained temporal attention map $Attn$ from the source video, we employ segment anything model ($SAM$) to obtain the mask of the moving objects in each frame $M_i=SAM(v_i), i=1,\cdots,t$, where $v_i$ denotes the $i$-th frame of \yr{the source video} $v_s$. 
Then, we \yr{merge} the masks of $t$ frames into one mask $M=U(M_1,M_2\cdots M_t)$. 
Since the \yr{motion in} the background \yr{region} mainly comes from the camera motion, we regard the masked \yr{temporal} attention map in the background region $Attn_m=Attn\odot (1-M)$ 
as the camera \yr{motion} attention map that only controls the camera motion. 
\yr{Although currently} the masked attention map $Attn_m$ has no 
\yr{value} 
\yr{inside} the \yr{moving objects} mask \yr{$M$}\yr{, we can estimate the camera motion inside the mask based on the camera motion outside}. 
\hut{To estimate the camera motion inside the mask \yr{$M$}, we transform the motion estimation problem into solving a Poisson equation, which is introduced below.}

\textbf{Predicting camera motion in foreground region.}
Video \yr{processing} tasks such as video compression, optical flow estimation, and video interpolation, 
share a common assumption that the changes between video frames are smooth and continuous~\cite{zhou2023mvflow,fleet2006optical,zhang2023modeling,gharibi2023multi}, and the motion\yr{s} of the pixels in a local neighborhood are similar. 
\yr{Based on} this assumption, we posit that the \textbf{camera motion} is also \textbf{continuous and has local coherence}\hut{, \yr{\textit{i.e.,}} the camera motions in a local region are almost the same}. 
Therefore, we assume the gradient \yr{of} the \yr{camera motion} attention map 
\yr{inside the mask region}
is quite small\yr{,} and the values of the attention map on both sides of the mask boundary are almost the same.
Denote the \yr{camera motion} attention map inside the mask \yr{$M$} as $A_{in}$ \yr{(to be estimated)}, and the \yr{camera motion} attention map outside the mask as $A_{out}$ \yr{(which we already have $A_{out}=Attn_m$)}. 
\yr{And we} denote the position\yr{s} of each pixel inside the mask as $\Omega\in\mathcal{R}^2$, 
\yr{and} the \yr{mask} boundary 
as $\partial \Omega$. 
Then, we have $\nabla A_{in} \approx 0$, and $A_{in}|_{\partial \Omega}=A_{out}|_{\partial \Omega}$. 
Since we already \yr{know $A_{out}$}, 
we can estimate $A_{in}$ by solving the following optimization problem:
\begin{equation}
\begin{aligned}
&A_{in}^*=\mathop{argmin}\limits_{A_{in}} \int\int_{\Omega} \|\nabla A_{in}\|^2. \\
&s.t.\, A_{in}|_{\partial \Omega}=A_{out}|_{\partial \Omega}.
\end{aligned}
\label{eq: camera motion estimation}
\end{equation}

Therefore, the camera-motion estimation problem is \yr{converted} into a Poisson blending problem. 
By setting the gradient inside the mask to be $0$, we can employ Successive Over Relaxation algorithm~\cite{young1954iterative} for Poission Blending to find 
the optimal solution $A_{in}^*$. 
Finally, we \yr{obtain} the complete camera \yr{motion} attention map $Attn^c=\{A_{in}\yr{^*},A_{out}\}$, which is disentangled with the object motion. 
With \yr{the disentangled} $Attn^c$, we can employ the camera motion transfer method in Sec.~\ref{sec:cmt} to transfer the camera motion from a single source video to target videos.

\subsection{Few-shot Camera Motion Disentanglement}
\label{sec:few_cme}
\yr{When the object motions are overly complex to disentangle, \textit{e.g.,} moving objects may occupy nearly all the pixels, it may be difficult to disentangle camera motion and object motion from a single video.}
To improve the disentanglement performance for videos with complex object motions, we relax the \yr{input} conditions from one shot to few shot.
\yr{I.e.,} we aim to extract the common camera motion from several videos $\{v_1, v_2\cdots v_m\}$ with similar camera motions.

\textbf{Extracting common feature in temporal attention as common camera motion.}
In Sec.~\ref{sec:one_cme}, we 
\yr{decompose}
the temporal attention maps \yr{of a single video} 
into camera motion and object motion. 
Since the given \yr{$m$} videos $\{v_1, v_2\cdots v_m\}$ share similar camera motions, we regard the common feature of the \yr{temporal} attention maps as the feature of camera motion. 
Therefore, \yr{we calculate common camera motion by extracting} a common feature \yr{from} the temporal attention map\yr{s of $m$ videos}. 
Since the motion at different locations may be different (\textit{e.g.,} zoom in/out), we model the motion at pixel level.
Denote the temporal attention map of each video as $\{A_1, A_2\cdots A_m\}$, where $A_i\in\mathcal{R}^{\yr{W}\times \yr{H}\times t\times t}$ and $t$ is the number of frames. 
For each pixel $(x,y)$ in video $v_i$, we denote its motion as $A_i(x,y)\in\mathcal{R}^{\yr{t \times t}}$.
\yr{Next, we aim to extract the common feature for each pixel $(x,y)$ from $m$ temporal attention maps.}

\textbf{Local coherence assumption for camera motion.}
\yr{To extract the common feature for each pixel $(x,y)$, only using the attention values at the location $(x,y)$ in $m$ temporal attention maps may not be adequate, especially when the object motions in the given $m$ video are complex and diverse.}
Therefore, based on the assumption of local coherence, we regard that the neighboring pixels \yr{in a window centered at pixel $(x,y)$} 
share similar camera motion \yr{as the center pixel}. 
In other words, \yr{we extract the \yr{common} camera motion for the pixel $(x,y)$ by} considering the attention values of neighboring pixels in \yr{a} $k\times k$ window $\mathcal{N}_k(x,y)$ 
\yr{in each of the $m$ temporal attention maps ($m\times k^2$ pixels in total), whose attention values form a tensor} 
$\mathcal{A}(x,y)=\{A_i(\mathcal{N}_k(x,y)),\yr{i=1\cdots m} \}\yr{\in R^{m\times k^2\times t\times t}}$.

\textbf{Extracting common camera motion by window-based clustering.}
For each pixel $(x,y)$, to extract the common camera motion from the 
attention values $\mathcal{A}(x,y)$ \yr{in its $k\times k$ neighboring window}, 
we first reshape the 
attention \yr{values $\mathcal{A}(x,y)$} 
\yr{to $R^{(m\times k^2)\times (t\times t)}$.}
We \yr{then} employ \yr{t}-SNE~\cite{van2008tsne} to reduce the dimension from $(t\times t)$ to $2$\yr{, for better clustering in the subsequent steps}. 
After dimension reduction, we 
\yr{compute}
the centroid of the \yr{$m\times k^2$ pixels} 
\yr{as the representation of} the \yr{common} camera motion. 
Directly computing the mean value of all the \yr{$m\times k^2$ pixels} is a possible solution to compute the centroid, but has inferior accuracy of the extracted motion when the camera motions in some of the samples are severely entangled with object motion. 
Therefore, we employ DBSCAN~\cite{ester1996dbscan} to cluster all the \yr{pixels}, 
which can effectively distinguish the outliers. 
After clustering, we have $n_c$ clusters, with each cluster containing part of the attention values. 
We regard the centroid of the largest cluster as the common camera motion, since it is the most common motion among 
\yr{the $m\times k^2$ pixels.} 
\yr{With} the extracted camera motion map 
$Attn^c$, 
we can transfer the camera motions to new videos.

\subsection{Camera Motion \yr{Combination}
}
\label{sec:cmb}

\textbf{Camera motion combination.}
In 
\yr{Sec.~\ref{sec:one_cme} and ~\ref{sec:few_cme}},
we extract the camera motion $Attn^c$ from a single or several videos. These camera motions can work separately by transferring one extracted camera motion to a target video. 
\yr{One natural question is}
whether we can combine different camera motions to enable a more complex and flexible camera motion control. 
\hut{To achieve this, \yr{in this section,} we \yr{explore} different ways \yr{to combine} camera motions, which enables 
\textbf{1)} combining different camera motions into a new motion;
\textbf{2)} applying different camera motions in different areas\yr{;} and \textbf{3)} preserving part of the contents while transferring the camera motion.}

\hut{\textbf{Additivity of the camera motions.} 
We first explore \yr{how} to combine different camera motions together. 
We are delighted to discover that the camera motions extracted from Sec.~\ref{sec:one_cme} and \ref{sec:few_cme} are additive. 
By \yr{adding} the attention maps $\{Attn_i^c\}_{i=1}^n$ corresponding to different camera motions, we can obtain a new camera motion that includes all the combined camera motions at the same time. 
And by assigning different weights $\{w_i\}_{i=1}^n$ to different camera motions, we can control the intensity of each camera motion by:}
\begin{equation}
\begin{aligned}
    Attn^c_{new}=\sum\limits_{i\in Sub(\{1\cdots n\})} w_i \times Attn^c_{i},
\end{aligned}
\label{eq: additivity of camera motions}
\end{equation}
where $Sub(\{1\cdots n\})$ is \yr{an} arbitrary subset of $\{1\cdots n\}$.


\textbf{Position-specified motion transfer.}
The camera motion transfer methods in previous sections can only transfer the camera motions in a global manner\yr{,} while lacking the ability to transfer the camera motions in a local \yr{region}.
Therefore, to enable our model with the ability to control the camera motions in a local manner, we propose a segment\yr{ation}-based local camera motion control method. 
\yr{We segment local regions by SAM, and}
assign different camera motions to different \yr{local regions} of the generated video, by applying the mask \yr{$M_{i}$} on the camera motion attention map \yr{$Attn^c_{i}$} as follows:
\begin{equation}
\begin{aligned}
    Attn^c_{new}=\sum\limits_i M_{i}\odot Attn^c_{i}.
\end{aligned}
\label{eq: camera motion basis}
\end{equation}

\begin{figure*}[t]
\centering
\includegraphics[width=1.0\textwidth]{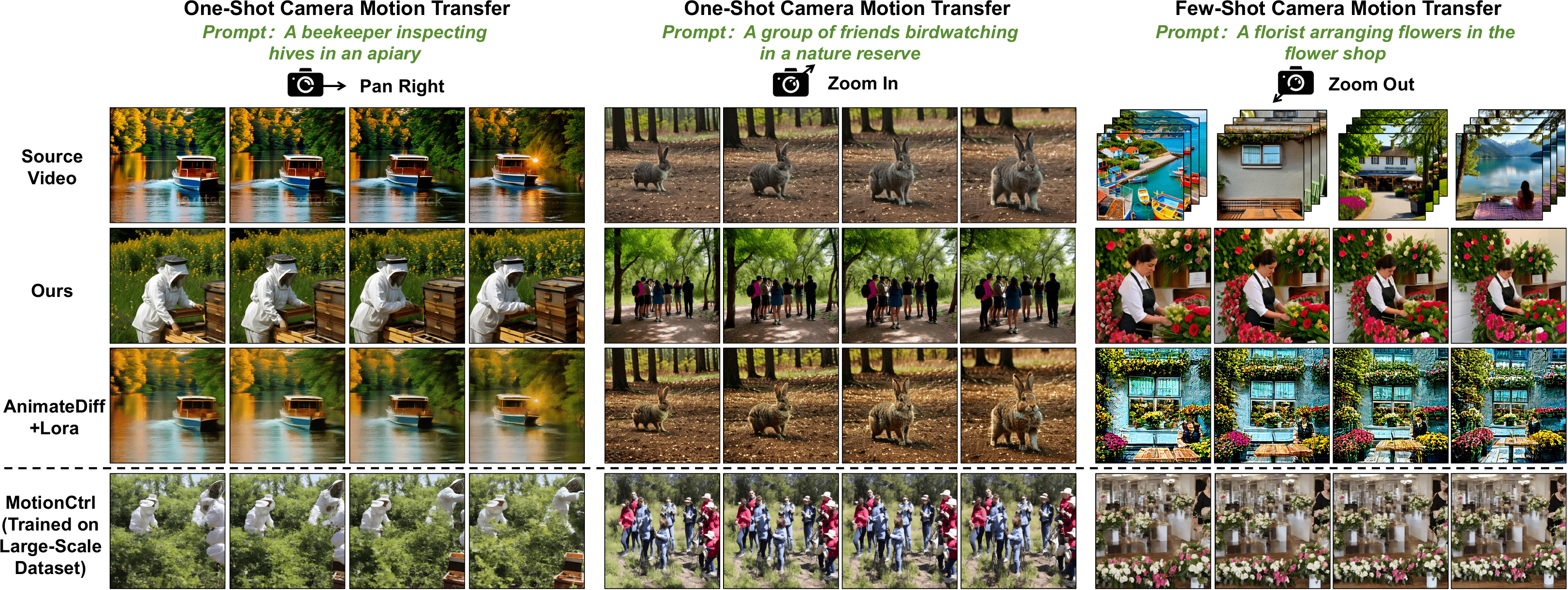}
\vspace{-0.2in}
\caption{The comparison on one-shot and few-shot camera motion transfer with AnimateDiff+Lora~\cite{guo2023animatediff} and MotionCtrl~\cite{wang2023motionctrl}. AnimateDiff+Lora tends to overfit to the training data while MotionCtrl suffers from shape distortions and logical inconsistencies when controlling camera motion, even though it is trained on large-scale data. In contrast, our MotionMaster generates high-quality videos with accurate camera motions.}
\label{fig:comparison on one-shot and few-shot}
\vspace{-0.15in}
\end{figure*}

\textbf{Local content-preserving camera motion transfer.} 
To better preserve specific content within the target video, we \yr{first} utilize 
\yr{SAM} to segment the object \yr{region} $M$ we aim to keep unchanged \yr{and then modify the temporal attention calculation}. 
We find that \yr{in diffusion-based video generation models}, 
the appearance and motions \yr{are} well \yr{disentangled} in the temporal attention modules, where the temporal attention maps represent the temporal motions, while the Value $V$ represents the appearance. 
Therefore, when we \yr{need} to transfer the camera motions from \yr{a} source video $v_s$ to a target video $v_t$ \yr{while} keep\yr{ing} the \yr{appearance in region} $M$ \yr{of} $v_t$ unchanged, 
\hut{we \yr{modify the temporal attention calculation by} keep\yr{ing} the Value inside \yr{$M$} the same as the Value $V_t$ of the target video, and substitut\yr{ing} the temporal attention map by the camera motion \yr{attention map} $Attn_s^c$ \yr{of} the source video, which can be \yr{formulated as follows}:}
\begin{equation}
\begin{aligned}
    V'=V_t\odot M+V\odot (1-M), ~~~
    ~~f_{out}= Attn^c_s V'.
\end{aligned}
\label{eq: content-preserving camera motion transfer}
\end{equation}

%% file: secs/4_experiment.tex
\begin{figure*}[t]
\centering
\includegraphics[width=0.75\textwidth]{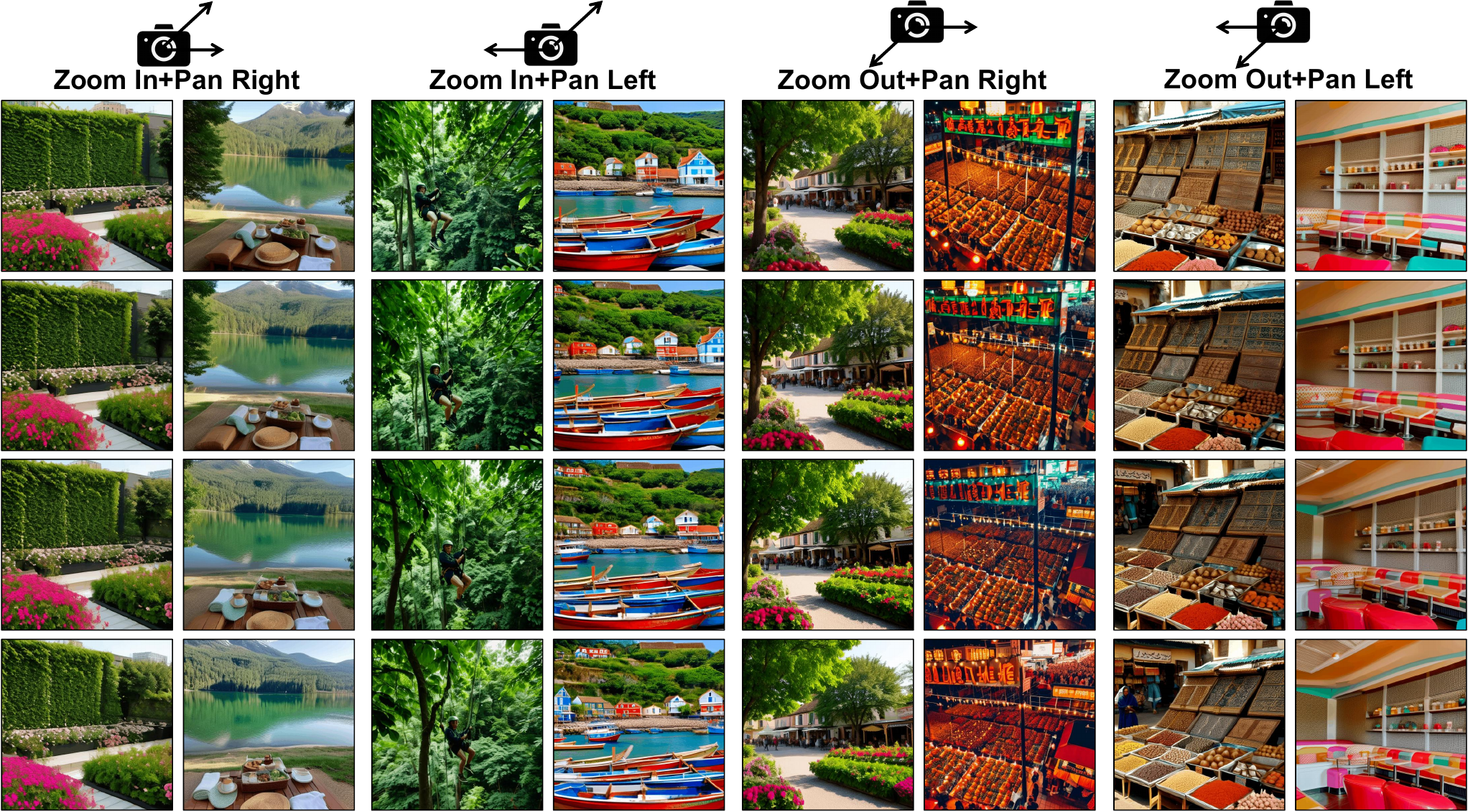}
\vspace{-0.05in}
\caption{\textbf{Camera motion combination results:} The extracted camera motions can be combined to form new camera motions. The newly constructed camera motions in this figure contain both zoom and pan camera motions at the same time.}
\label{fig:results of combing camera motions}
\vspace{-0.1in}
\end{figure*}

\section{Experiments}
\subsection{Implementation Details}
\textbf{Experiment details and hyperparameters.} 
\yyy{In our experiments, we adopt AnimateDiff~\cite{guo2023animatediff} as the baseline method for motion disentanglement and control,}
which is one of the state-of-the-art text-to-video models. 
The generated video size is $512\times 512$, with each video composed of 16 frames with 8 \yyy{FPS}. 
When generating videos, we employ 25-step DDIM~\cite{song2020ddim} for inference and choose the temporal attention maps in the 15-th step to extract the camera motions. Moreover, for few-shot camera motion extraction, we compute the neighborhood size $k$ by $k=\lceil\frac{size}{16}\rceil\times2+1$, where $size$ is the width and height of the temporal attention maps.

\textbf{Evaluation \yyy{m}etrics.} 
To evaluate the generation quality, diversity and camera motion accuracy, we employ \yyy{three evaluation metrics:}
1) FVD~\cite{unterthiner2018towards}: Fr\'echet Video Distance 
measures \yyy{the} quality and authenticity by calculating the Fr\'echet distance between real and generated videos; 2) FID-V~\cite{balaji2019conditional}: Video-level FID uses a 3D Resnet-50 model to extract video features for video-level FID scoring, measuring the quality and diversity of the generated videos; and 3) Optical Flow Distance~\cite{farneback2003two} assesses \yyy{the} camera movement accuracy by computing distance between the flow maps from \yyy{the} generated and ground truth videos. 

\subsection{Camera Motion Transfer}
\label{sec:exp-camera-motion-transfer}

\textbf{Qualitative comparison with the state-of-the-arts.} 
To 
\yyy{validate} the effectiveness of our model, we compare our model with the state-of-the-art camera motion control methods on four types of basic camera motions: 1) zoom in, 2) zoom out, 3) pan left, and 4) pan right (in \#Suppl). 
\yyy{We compare with two motion control methods: 1)} AnimateDiff~\cite{guo2023animatediff} employs the temporal LoRA~\cite{hu2021lora} module to learn the motions from given videos with target camera motions. We train motion LoRA modules on AnimateDiff with one-shot and few-shot data, and compare them with our model. 
\yyy{2)} Moreover, we also compare with MotionCtrl~\cite{wang2023motionctrl}. Since the training \yyy{code} is not open-sourced, we employ the \yyy{officially provided} model\yyy{, which is} pretrained on a large scale of camera-labeled data. 

The comparison results are shown in Fig.~\ref{fig:comparison on one-shot and few-shot} \yyy{(video comparison results are provided in \#Suppl)}. 
It can be seen that in one-shot condition, Animate\yyy{D}iff tends to overfit to the given video\yyy{;} while in the few-shot condition, AnimateDiff tends to mix the features of the training videos, which cannot generate correct videos corresponding to the given prompts. 
Motion\yyy{C}trl can generate videos that better align with the prompts, but may cause shape distortions and logical inconsistencies when 
controlling camera motion. 
In contrast, our model can generate high-quality and diverse videos with only one-shot or few-shot data\yyy{,} without the need for training. 

\textbf{Quantitative comparison.} We also compare \yyy{with} these models quantitatively\yyy{, using} FVD, FID-V, and Optical Flow distance to evaluate the generation quality, diversity, and camera motion accuracy. For each method, we generate 1,000 videos for each type of camera motion and compute FVD and FID-V with 1,000 collected high-quality videos. We also compute the average Optical Flow Distance between the generated videos and given videos. The results are shown in Tab.~\ref{tab:quantitative comparison}, where our model achieves the best FID-V and FVD, demonstrating superior generation quality and diversity. Since Animate\yyy{D}iff overfits to the training data, it get a lower Flow distance, but suffers from the worst generation diversity. In summary, our model achieves the best FVD and F\yyy{ID}-V, while also ensuring a good camera transfer accuracy compared to MotionCtrl.


\begin{table}[t]
\centering
\caption{Quantitative comparison results with the state-of-the-art methods on FVD, FID-V and Optical Flow Distance. 
Note that AnimateDiff+Lora~\cite{guo2023animatediff} overfits to the training data, thereby achieving the lowest flow distance. But FVD and FID-V demonstrate its worst generation diversity. In contrast, our model achieves the best FVD and F\yyy{ID}-V, while also ensuring a good camera transfer accuracy compared to MotionCtrl~\cite{wang2023motionctrl}.
}
\vspace{-0.05in}
\begin{tabular}{c}
\begin{minipage}[t]{1.0\linewidth}
\resizebox{1.0\linewidth}{!}{
\begin{tabular}{cc|ccc|ccc}
\toprule
\multicolumn{2}{c}{Data and Method} & \multicolumn{3}{|c|}{Pan Right} & \multicolumn{3}{c}{Zoom In} \\ 
Data Scale & Method & FID-V $\downarrow$ & FVD $\downarrow$& Flow Dis $\downarrow$ & FID-V $\downarrow$& FVD $\downarrow$& Flow Dis $\downarrow$\\
\midrule
\multirow{2}{*}{One shot} & AnimateDiff & 382.40 & 4956.42 & \textbf{19.76} & 482.58 & 6322.46 & \textbf{6.91} \\
 & \textbf{MotionMaster} & \textbf{54.45} & \textbf{921.95} & \underline{37.92} & \textbf{61.45} & \textbf{863.24} & \underline{12.11} \\ \midrule
Large Scale & MotionCtrl & 95.83 & 1207.52 & 38.18 & 80.58 & 935.08 & 13.12\\
\bottomrule
\end{tabular}}
\caption*{(a) Comparison results on one-shot camera motion control.}
\vspace{-0.1in}
\end{minipage}
\end{tabular}

\begin{tabular}{c}
\begin{minipage}[t]{1.0\linewidth}

\resizebox{1.0\linewidth}{!}{
\begin{tabular}{cc|ccc|ccc}
\toprule
\multicolumn{2}{c}{Data and Method} & \multicolumn{3}{|c|}{Pan Left} & \multicolumn{3}{c}{Zoom Out} \\ 
Data Scale & Method & FID-V $\downarrow$& FVD $\downarrow$& Flow Dis $\downarrow$& FID-V $\downarrow$& FVD $\downarrow$& Flow Dis $\downarrow$\\
\midrule
\multirow{2}{*}{Few shot} & AnimateDiff & 268.29 & 4629.08 & \textbf{14.76} & 251.44 & 3975.41 & \textbf{3.12} \\
 & \textbf{MotionMaster} & \textbf{61.38} & \textbf{1092.09} & \underline{38.94} & \textbf{52.90} & \textbf{910.76} & \underline{5.10} \\ \midrule
Large Scale & MotionCtrl & 98.04 & 1196.54 & 55.25 & 80.12 & 928.41 & 7.88\\
\bottomrule
\end{tabular}}
\caption*{(b) Comparison results on few-shot camera motion control.}

\end{minipage}
\end{tabular}

\label{tab:quantitative comparison}
\vspace{-0.3in}
\end{table}

\subsection{Flexible Motion Control}

\textbf{Motion combination.}
In this section, we evaluate the additivity of our disentangled camera motion attention maps. We employ the extracted camera motions including zoom in, zoom out, pan left and pan right in Sec.~\ref{sec:exp-camera-motion-transfer} and combine two of them into a new camera motion by Eq.(\ref{eq: additivity of camera motions}). The results are shown in Fig.~\ref{fig:results of combing camera motions}. It can be seen that when combining the zooming motions with the panning motions, the camera zooms and pans at the same time, which demonstrates that our model can successfully combine different kinds of camera motions together while ensuring generation quality.


\begin{figure}[t]
\centering
\includegraphics[width=0.45\textwidth]{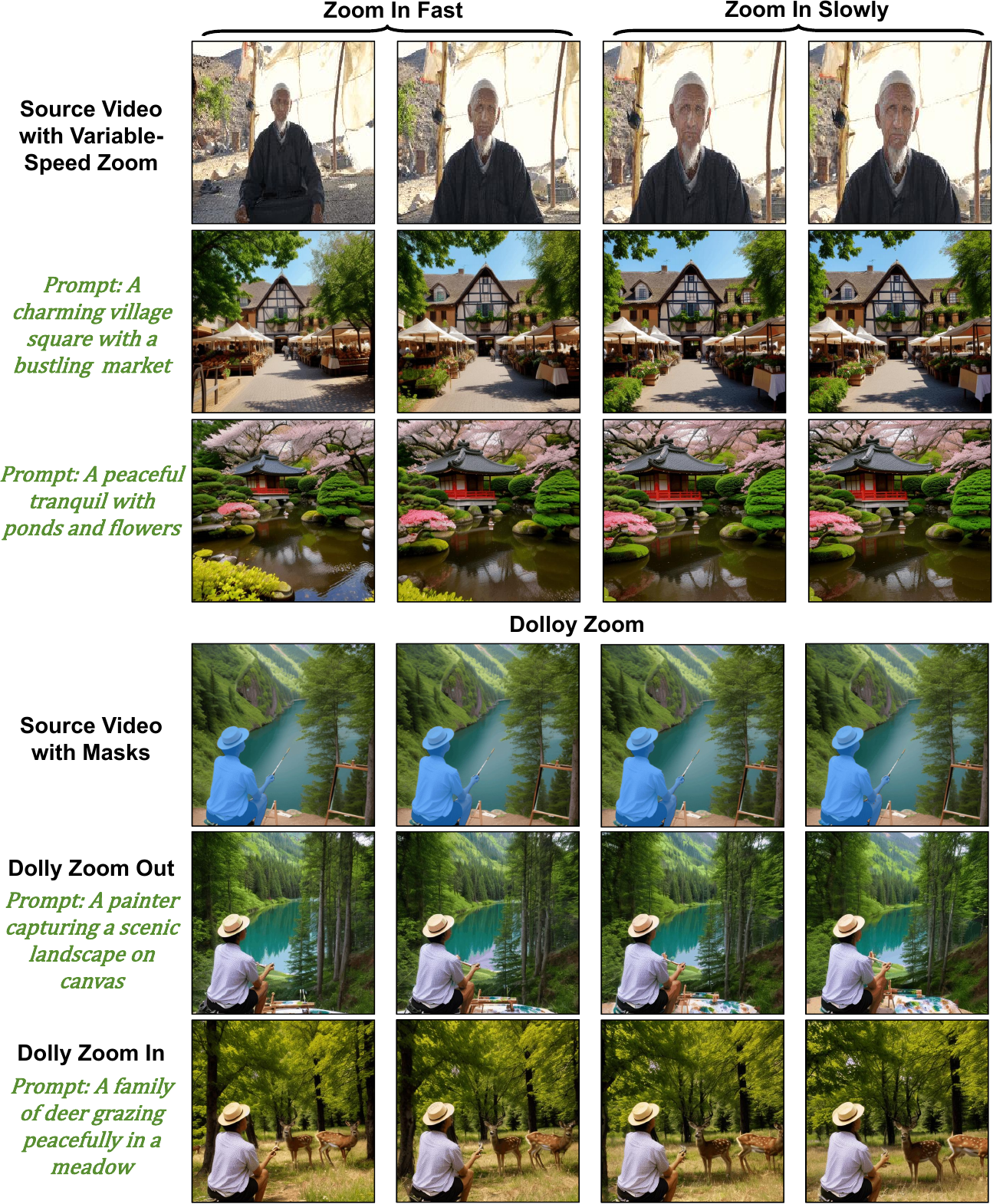}
\vspace{-0.1in}
\caption{Camera motion control results on professional camera motions, including variable-speed zoom and dolly zoom.}
\label{fig:sudden zoom and dolly zoom}
\vspace{-0.1in}
\end{figure}

\textbf{More professional camera motions.} 
In this section, we show more professional camera motions in the real film industry, including variable-speed zoom and dolly zoom. 
\yyy{For} variable-speed zoom, where the camera firstly zooms in fast and then zooms in slowly, we crop a video \yyy{clip} from films with \yyy{this kind of} motion, \yyy{and achieve this motion control by one-shot camera motion disentanglement (Sec.~\ref{sec:one_cme})}. 
For dolly zoom\yyy{,} where the camera in the background region zooms while the camera in the foreground fixes, we employ the local content-preserving camera motion transfer method (Sec.~\ref{sec:cmb}) to realize it. 
The results are shown in Fig.~\ref{fig:sudden zoom and dolly zoom}. 
It can be seen that our model transfers the variable-speed zoom motion in the given video well, and achieves good generation results in both dolly zoom in and \yyy{dolly zoom} out motion controls.

\begin{figure}[t]
\centering
\includegraphics[width=0.45\textwidth]{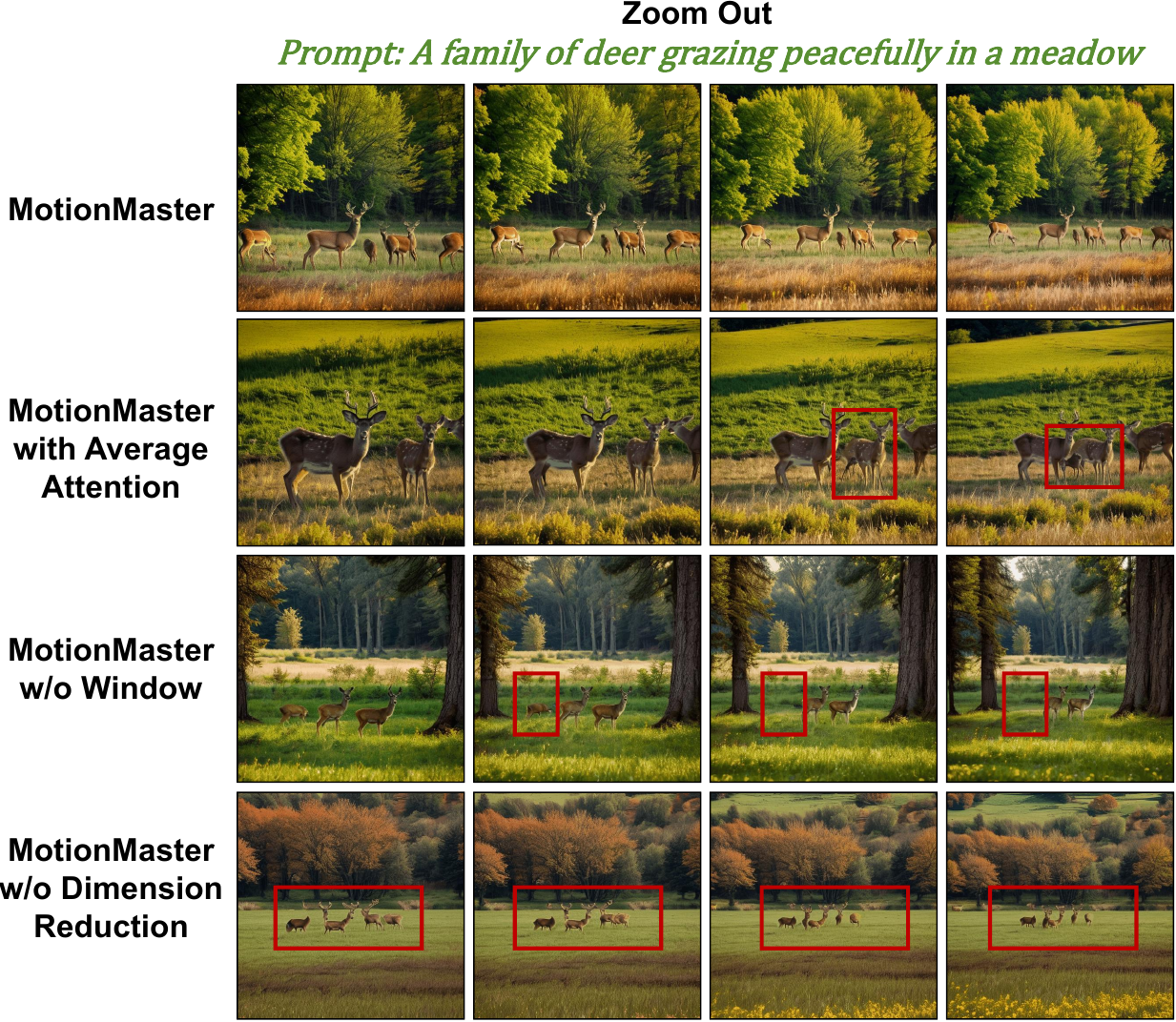}
\vspace{-0.05in}
\caption{Ablation study on one-shot camera motion disentanglement. The model without motion disentanglement generated artifacts in the region of the moving rabbit.}
\label{fig:ablation study on one shot}
\vspace{-0.1in}
\end{figure}

\begin{figure}[t]
\centering
\includegraphics[width=0.45\textwidth]{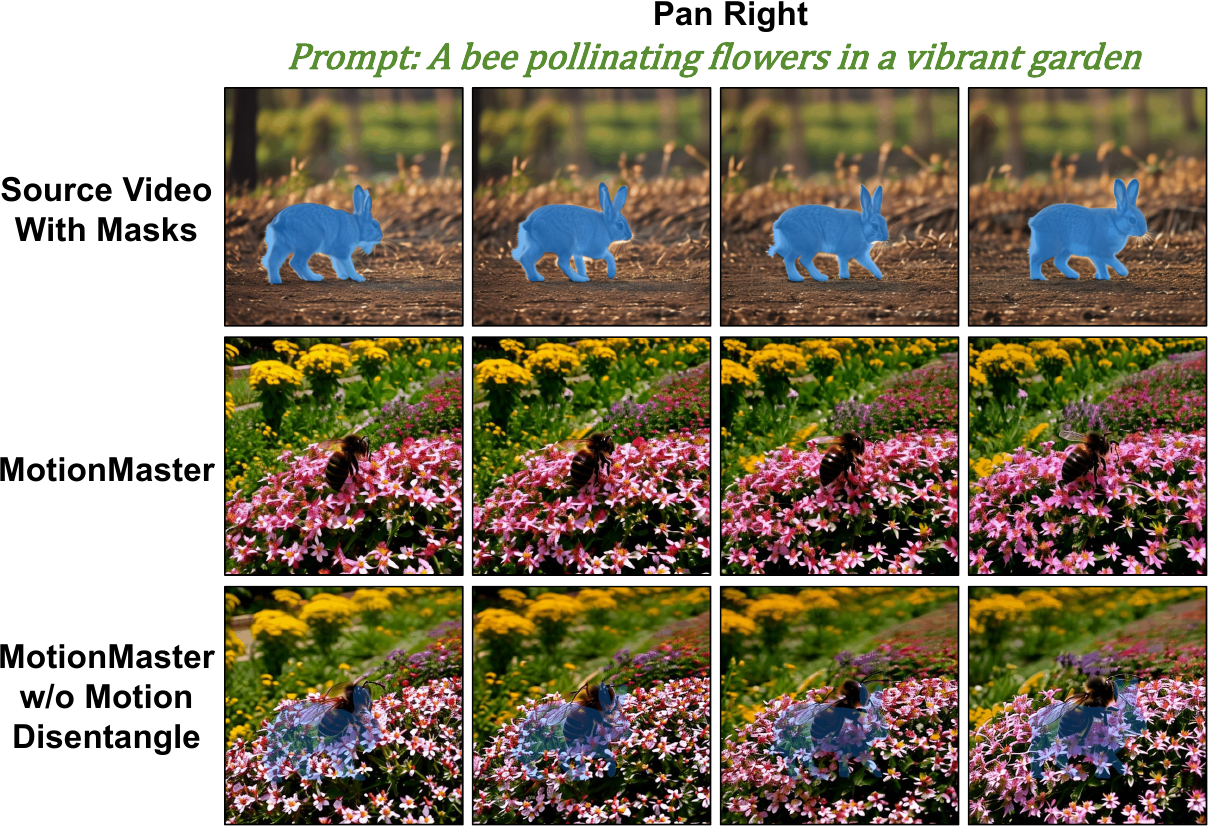}
\vspace{-0.05in}
\caption{Ablation study on few-shot camera motion disentanglement. All the ablated models generate videos with unnatural movements shown in the red boxes which are caused by the inaccurate extracted camera motions.}
\label{fig:ablation study on few shot}
\vspace{-0.2in}
\end{figure}


\subsection{Ablation Study}
\textbf{Ablation on one-shot camera motion disentanglement.} We first \yyy{validate} the effectiveness of our one-shot camera motion disentanglement method. We compare our model with the \yyy{ablated version} that directly transfers the temporal attention map from the source video to the target video, which does not disentangle the camera and object motions. The results are shown in Fig.~\ref{fig:ablation study on one shot}. It can be seen that when transferring the pan right camera motion entangled with the \yyy{object} motion of the moving rabbit, the model without motion disentanglement tends to generate artifacts in the region of the rabbit, which is clearer in the video of \#Suppl.

\textbf{Ablation on few-shot camera motion disentanglement.} We then \yyy{validate} the effectiveness of our few-shot camera motion disentanglement method. We experiment on three ablated versions on zoom-out camera motion: 1) \yyy{MotionMaster with Average Attention:} the model without DBSCAN \yyy{clustering} and \yyy{directly} averages the camera motions from all the videos; 
2) \yyy{MotionMaster w/o Window:} the model without the window-based clustering\yyy{,} which only uses \yyy{the} $m$ \yyy{pixels at the same location} 
for clustering; 
and 3) \yyy{MotionMaster w/o Dimension Reduction:} the model without \yyy{t}-SNE to reduce the dimension. 
The comparison results are shown in Fig.~\ref{fig:ablation study on few shot}. 
It can be seen that all the ablated models generate unnatural movements shown in the red boxes where certain objects abruptly appear or vanish, or \yyy{suffer from} shape distortions. 
In contrast, our model achieves the highest generation quality and transfers the camera motions correctly.


%% file: secs/5_conclusion.tex
\section{Conclusion}
In this paper, we propose MotionMaster, a training-free camera motion transfer method based on camera-motion disentanglement. 
We \yyy{find} that the temporal attention map in the video diffusion model is composed of both camera motion and object motion. We then propose two methods to disentangle the camera motions from object motions for a single or several videos. Moreover, with the extracted camera motions, we further propose a camera motion combination method to enable our model a more flexible and controllable camera control. Extensive experiments demonstrate the superior camera motion transfer ability of our model and show \yyy{our} great potential in controllable video generation.

%% file: secs/6_suppl.tex
\section{Overview}

\yr{In this supplementary material, more details about the proposed MotionMaster and more experimental results are provided, including:}

\begin{itemize}
    \item More implementation details (Sec.~\ref{sec:More implementation details});
    \item Solving Poisson Equation (Sec.~\ref{sec:Solving Poisson Equation});
    \item Temporal attention maps determines the video motion (Sec.~\ref{sec:Temporal Attention Maps Determines The Video Motion})
    \item More Comparison Results (Sec.~\ref{sec:More Comparison Results});
    \item More Experiments on the hyperparameters (Sec.~\ref{sec:More Experiments on the hyperparameters});
    \item User Study (Sec.~\ref{sec:User Study}).
\end{itemize}

To see the generated results more clearly, you can refer to
\url{https://sjtuplayer.github.io/projects/MotionMaster}, which includes all the videos in the experiments.

\section{More Implementation Details}
\label{sec:More implementation details}
We conduct experiments based on AnimateDiff-v2~\cite{guo2023animatediff}. we use DDIM~\cite{song2020denoising} to accelerate the generation process with 25 denoising steps. Moreover, to decrease the computation cost, we employ the temporal attention maps in timestep $t=15$ to represent the video motions in different timesteps as illustrated in Sec. 3.2 of the main paper. Furthermore, for few-shot camera motion disentanglement, we specified a video count of 5 and configured DBSCAN clustering~\cite{ester1996dbscan} with an Eps-neighborhood of 4 and core points of 3.

\section{Solving Poisson Equation}
\label{sec:Solving Poisson Equation}

We complete the temporal attention map inside the object-moving region by solving a Poisson equation. The gradients within object-moving regions of the completed attention map are assumed to be zero and the boundary values should match those of the original attention map.
We choose the parallel red-black ordering Gauss-Seidel iteration method to solve the Poisson equation. Initially, we label the pixels with red-black ordering, ensuring that each pixel and its neighboring pixels alternate between being labeled red and black. Next, while ensuring that the values of boundary nodes remain unchanged, we update the red and black pixels alternately until reaching a specified number of iterations or until the residual falls below a predefined threshold. The iteration process is illustrated by the pseudo code. This algorithm can be accelerated using parallel computing frameworks like CUDA.

\section{Temporal Attention Maps Determines The Video Motion}
\label{sec:Temporal Attention Maps Determines The Video Motion}
\begin{figure*}[t]
\centering
\includegraphics[width=1.0\textwidth]{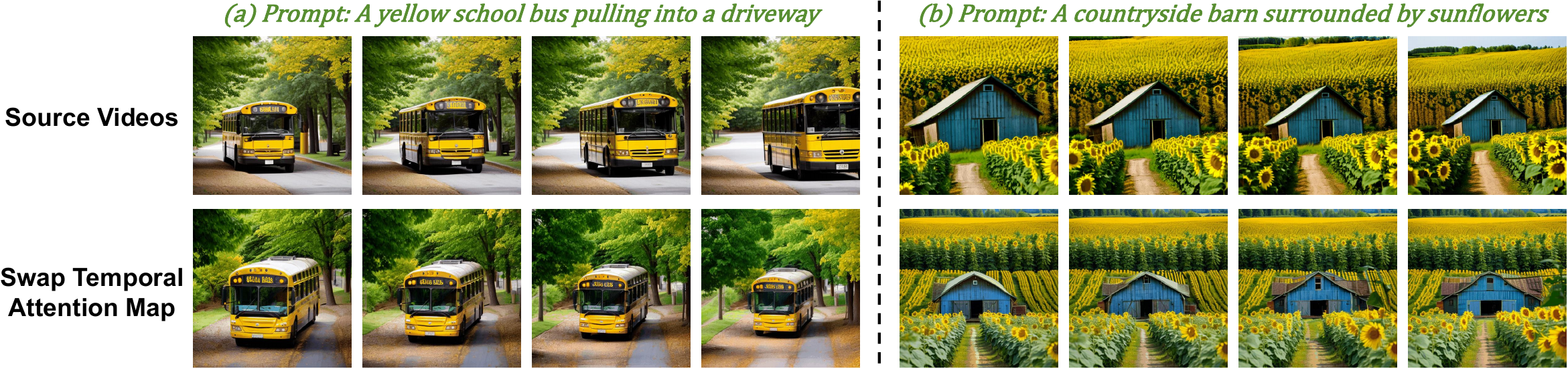}
\caption{After swapping the temporal attention maps of the first-row videos, we have the second-row videos which swap the motions. The source videos of (a) moves the bus while fixing the camera, and (b) keeps the objects fixed while zooming out the camera. After swapping the temporal attention map, (a) keeps the bus fixed while zooming out the camera. And (b) fixed camera, but there is a shadow of a moving bus in the generated video (it is clearer in demo.html).}
\label{fig:swapping the temporal attention maps}
\end{figure*}

The foundation of our method comes from the observation that the temporal attention map determines the motions in the generated videos, including camera motions and object motions. To validate this, we conduct an experiment to swap the temporal attention maps between two videos, where one of them contains only camera motion while the other one contains only object motion. The results are shown in Fig.~\ref{fig:swapping the temporal attention maps}. It can be seen that after swapping the temporal attention map, the contents of the two videos are similar and the motions are totally swapped. The source videos of (a) keep the camera fixed while moving the bus from left to right and (b) keep the object fixed while zooming out the camera. After swapping the temporal attention maps, the second row of (a) keeps the bus fixed while zooming out the camera and (b) keeps the camera fixed, but a shadow of a bus moves from left to right. Therefore, the temporal attention maps determine both the camera and object motions and by swapping the temporal attention map, the motions can be transferred to a new video.

\begin{algorithm}[t]
\caption{Solving Poisson Equation}
\label{alg:poisson equation}
\begin{algorithmic}[1]
\Function{poisson\_solving}{u, f}
    \State u: RGB, f: gradient
    \State choose an initial guess $u^{(0)}$
    
    \While {not converge}:
        \For {(i, j) is red node}:
        \State $u_{i,j}^{(k+1)}=\frac{1}{4}(f_{i,j}+u_{i+1,j}^{(k)}+u_{i-1,j}^{(k)}+u_{i,j+1}^{(k)}+u_{i,j-1}^{(k)})$
        \EndFor
        \For{(i,j) is black node}:
        \State $u_{i,j}^{(k+1)}=\frac{1}{4}(f_{i,j}+u_{i+1,j}^{(k+1)}+u_{i-1,j}^{(k+1)}+u_{i,j+1}^{(k+1)}+u_{i,j-1}^{(k+1)})$
        \EndFor
    \EndWhile
\EndFunction
\end{algorithmic}

\end{algorithm}

\begin{table*}[t]
\centering
\caption{Quantitative comparison results with the state-of-the-art methods on FVD, FID-V, and Optical Flow Distance. 
Note that AnimateDiff+Lora~\cite{guo2023animatediff} overfits to the training data, thereby achieving the lowest flow distance. But FVD and FID-V demonstrate its worst generation diversity. In contrast, our model achieves the best FVD and F\yyy{ID}-V, while also ensuring a good camera transfer accuracy compared to MotionCtrl~\cite{wang2023motionctrl}.
}
\begin{tabular}{c}
\begin{minipage}[t]{0.96\linewidth}
\resizebox{1.0\linewidth}{!}{
\begin{tabular}{cc|ccc|ccc|ccc|ccc}
\toprule
\multicolumn{2}{c}{Camera Motion} & \multicolumn{3}{|c}{Pan Right} & \multicolumn{3}{|c}{Pan Left} & \multicolumn{3}{|c}{Zoom In} & \multicolumn{3}{|c}{Zoom Out} \\
Data Scale & Method & FVD $\downarrow$ & FID-V $\downarrow$ & Flow Dis $\downarrow$  & FVD $\downarrow$ & FID-V $\downarrow$ & Flow Dis $\downarrow$ & FVD $\downarrow$ & FID-V $\downarrow$ & Flow Dis $\downarrow$ & FVD $\downarrow$ & FID-V $\downarrow$ & Flow Dis $\downarrow$ \\
\midrule
\multirow{2}{*}{One shot} & Animatediff & 382.4 & 4956.42 & \textbf{19.76} & 382.04 & 5939.96 & \textbf{15.22} & 482.58 & 6322.46 & \textbf{6.91} & 396.96 & 7767.33 & \textbf{5.78} \\
 & \textbf{MotionMaster (ours)} & \textbf{54.45} & \textbf{921.95} & \underline{ 37.92} & \textbf{64.43} & \textbf{933.77} & \underline{ 35.64} & \textbf{61.45} & \textbf{863.24} & \underline{ 12.11} & \textbf{55.23} & \textbf{862.9} & \underline{ 6.93} \\
Largescale & MotionCtrl & 95.83 & 1207.52 & 38.18 & 98.04 & 1196.54 & 55.25 & 80.58 & 935.08 & 13.12 & 80.12 & 928.41 & 7.88\\
\bottomrule
\end{tabular}}
\caption*{(a) Comparison results on one-shot camera motion control. \textbf{Bold} and \underline{underline} represent optimal and sub-optimal results, respectively.}
\vspace{-0.1in}
\end{minipage}
\\
\begin{minipage}[t]{0.96\linewidth}
\resizebox{1.0\linewidth}{!}{
\begin{tabular}{cc|ccc|ccc|ccc|ccc}
\toprule
\multicolumn{2}{c}{Camera Motion} & \multicolumn{3}{|c}{Pan Right} & \multicolumn{3}{|c}{Pan Left} & \multicolumn{3}{|c}{Zoom In} & \multicolumn{3}{|c}{Zoom Out} \\
Data Scale & Method & FVD $\downarrow$ & FID-V $\downarrow$ & Flow Dis $\downarrow$  & FVD $\downarrow$ & FID-V $\downarrow$ & Flow Dis $\downarrow$ & FVD $\downarrow$ & FID-V $\downarrow$ & Flow Dis $\downarrow$ & FVD $\downarrow$ & FID-V $\downarrow$ & Flow Dis $\downarrow$ \\
\midrule
\multirow{2}{*}{Few shot} & Animatediff & 290.86 & 5198.78 & \textbf{25.61} & 268.29 & 4629.08 & \textbf{14.76} & 281.73 & 4333.26 & \textbf{5.72} & 251.44 & 3975.41 & \textbf{3.12} \\
 & \textbf{MotionMaster (ours)} & \textbf{55.94} & \textbf{1153.27} & \underline{ 35.98} & \textbf{61.38} & \textbf{1092.09} & \underline{ 38.94} & \textbf{51.97} & \textbf{847.08} & \underline{ 12.93} & \textbf{52.90} & \textbf{910.76} & \underline{ 5.10} \\
Largescale & MotionCtrl & 95.83 & 1207.52 & 38.18 & 98.04 & 1196.54 & 55.25 & 80.58 & 935.08 & 13.12 & 80.12 & 928.41 & 7.88\\
\bottomrule
\end{tabular}}
\caption*{(b) Comparison results on few-shot camera motion control. \textbf{Bold} and \underline{underline} represent optimal and sub-optimal results, respectively.}
\end{minipage}

\end{tabular}
\label{tab:suppl_quantitative comparison}
\vspace{-0.15in}
\end{table*}

\section{More Comparison Results}
\label{sec:More Comparison Results}

\begin{figure*}[t]
\centering
\includegraphics[width=1.0\textwidth]{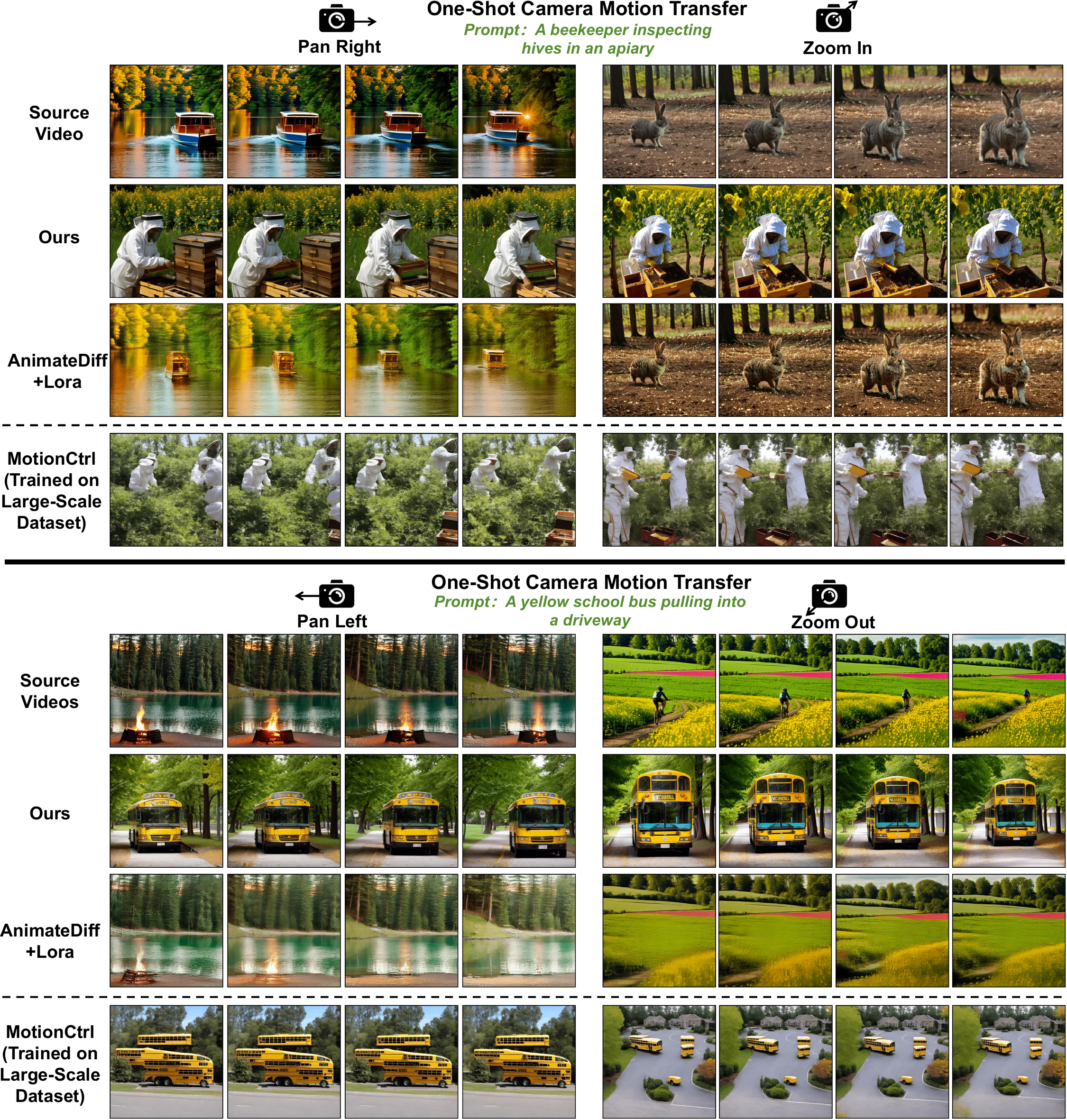}
\caption{Comparison on one-shot camera motion transfer on four basic camera motions: pan right, pan left, zoom in and zoom out. AnimateDiff+Lora~\cite{guo2023animatediff} overfits the training video. Even if MotionCtrl~\cite{wang2023motionctrl} is trained on a large-scale dataset, it still suffers from inaccurate camera motion control and some artifacts in the generated videos. In contrast, our model accurately transfers the camera motions while ensuring good generation quality and diversity. }
\label{fig:one-shot-comparison}
\end{figure*}

\begin{figure*}[t]
\centering
\includegraphics[width=1.0\textwidth]{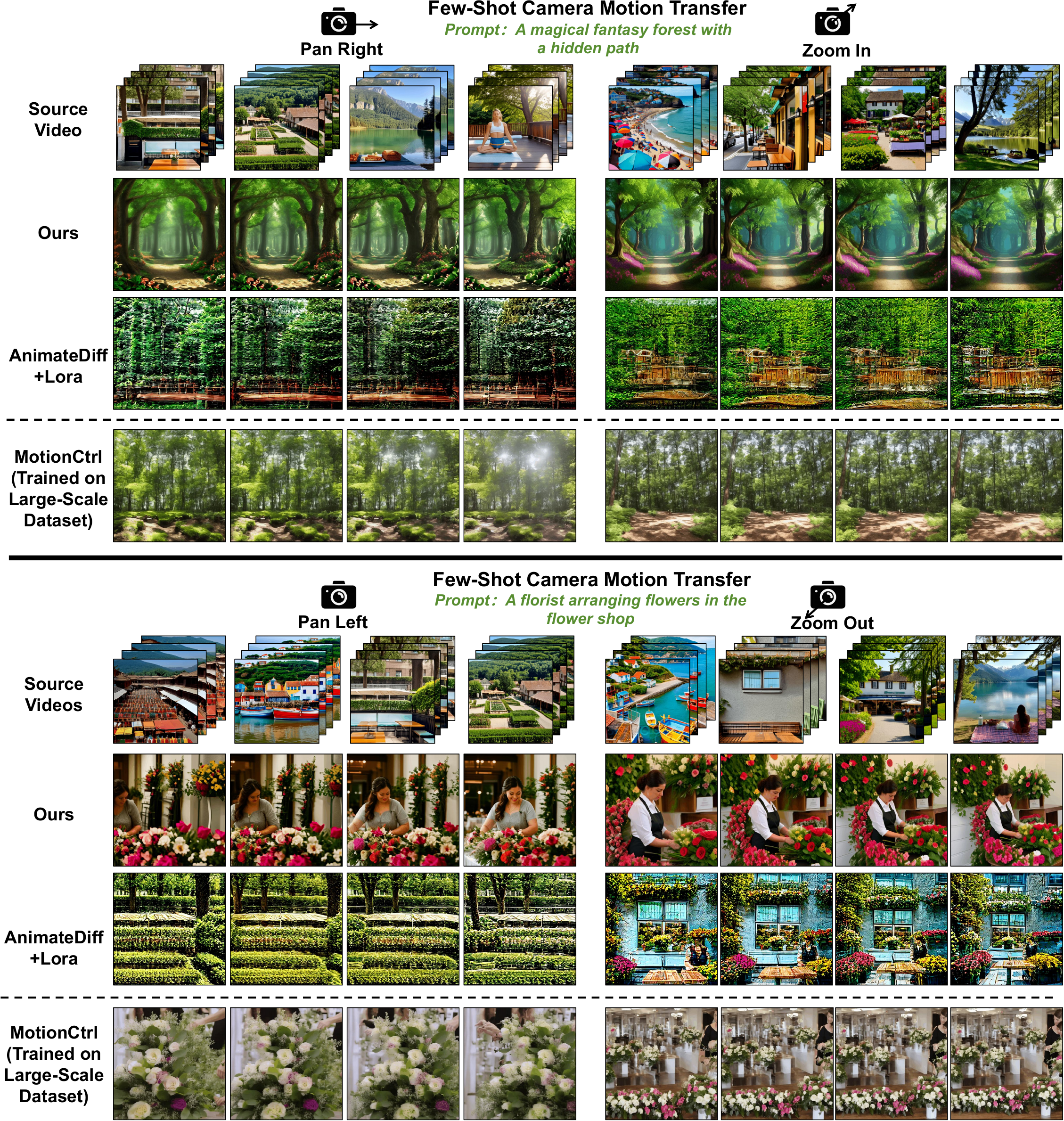}
\caption{Comparison on few-shot camera motion transfer on four basic camera motions: pan right, pan left, zoom in and zoom out. AnimateDiff+Lora~\cite{guo2023animatediff} overfits to the training videos, which generate videos with mixed features from the training data. Even if MotionCtrl~\cite{wang2023motionctrl} is trained on a large-scale dataset, it still suffers from inaccurate camera motion control and some artifacts in the generated videos. In contrast, our model accurately transferred the camera motions while ensuring good generation quality and diversity. }
\label{fig:few-shot-comparison}
\end{figure*}

\textbf{Qualitative comparison.} In this section, we show more results on one-shot and few-shot camera motion transfer results, where both one-shot and few-shot methods are employed to transfer zoom-in, zoom-out, pan-left, and pan-right camera motions. The qualitative comparison results are shown in Fig.~\ref{fig:one-shot-comparison} and \ref{fig:few-shot-comparison}. 
In the one-shot scenario, AnimateDiff+Lora~\cite{guo2023animatediff} appears prone to overfitting to the provided video, whereas in the few-shot scenario, it tends to amalgamate features from the training videos, leading to inaccurate video generation in response to the given prompts. MotionCtrl~\cite{wang2023motionctrl} exhibits improved alignment with prompts in video generation; however, it may introduce shape distortions and logical inconsistencies in camera motion control. In contrast, our model achieves high-quality and high-diversity generation with only one-shot or few-shot data, without the need for training. 

\textbf{Quantitative comparison.} To further validate the effectiveness of our model, we conduct quantitative comparisons on the four basic camera motions with one-shot and few-shot data. The comparison results are shown in Tab.~\ref{tab:suppl_quantitative comparison}. It shows that our model achieves the best FVD~\cite{unterthiner2018towards} and FID-V~\cite{balaji2019conditional} scores, indicating the best generation quality and diversity of our model. Since Animatediff is overfitted to the training data, it has the minimum optical flow distance~\cite{farneback2003two}, but it suffers from much worse FVD and FID-V. In summary, our model achieves the best FVD and FID-V, while also ensuring a good camera transfer accuracy compared to MotionCtrl. (Note that the header of Tab. 1(b) in the main paper should be "Pan Left" and "Zoom Out" and Tab.~\ref{tab:suppl_quantitative comparison} here is the correct version)

\textbf{Comparison on the computation cost.} Moreover, we also compare the computation cost including computation time and GPU memory requirement between MotionMaster, Animatediff+Lora~\cite{guo2023animatediff} and MotionCtrl~\cite{chen2023motion}. Since our model is a training-free method, we compute the time for disentangling the camera-object motions as our training time. To ensure the fairness of the experiment, we compute the time on the same NVIDIA A100 GPU. Meanwhile, we compare the GPU memory required for all the methods. The comparison results are shown in Tab.~\ref{tab:comparison on computation resources}. It can be seen that our model can accomplish camera motion disentanglement in a few minutes while the other methods require a much longer training time. Moreover, both AnimateDiff+Lora and MotionCtrl require more than 30G GPU memory, while our model only needs 13G GPU memory which is the only method that can be implemented on a single 24G 3090/4090 GPU.

\begin{table}[t]
\setlength{\abovecaptionskip}{4pt}
\setlength{\belowcaptionskip}{-0.2cm}
\setlength\tabcolsep{5pt}
\renewcommand{\arraystretch}{1.2}
\caption{Comparison on the computation resources. Our training-free MotionMaster requires much less time to control the camera motions and it is the only method that is capable of running on a single NVIDIA 24G 3090/4090 GPU.}
\resizebox{1.0\linewidth}{!}{
\begin{tabular}{c|cc}
\toprule
Method           & Computation Time & GPU Memory \\
\midrule
Animatediff+Lora~\cite{guo2023animatediff} & $\approx$10 hours       & 52G        \\
MotionCtrl~\cite{wang2023motionctrl}       & $>$ 10 days        & 32G        \\
One-shot MotionMaster (Ours)   & $\approx$ 60 seconds     & 13G        \\
Few-shot MotionMaster (Ours)   & $\approx$150 seconds    & 13G   \\
\bottomrule
\end{tabular}}
\label{tab:comparison on computation resources}
\end{table}

\section{More Experiments on The Hyperparameters}
\label{sec:More Experiments on the hyperparameters}

\begin{figure*}[t]
\centering
\includegraphics[width=1.0\textwidth]{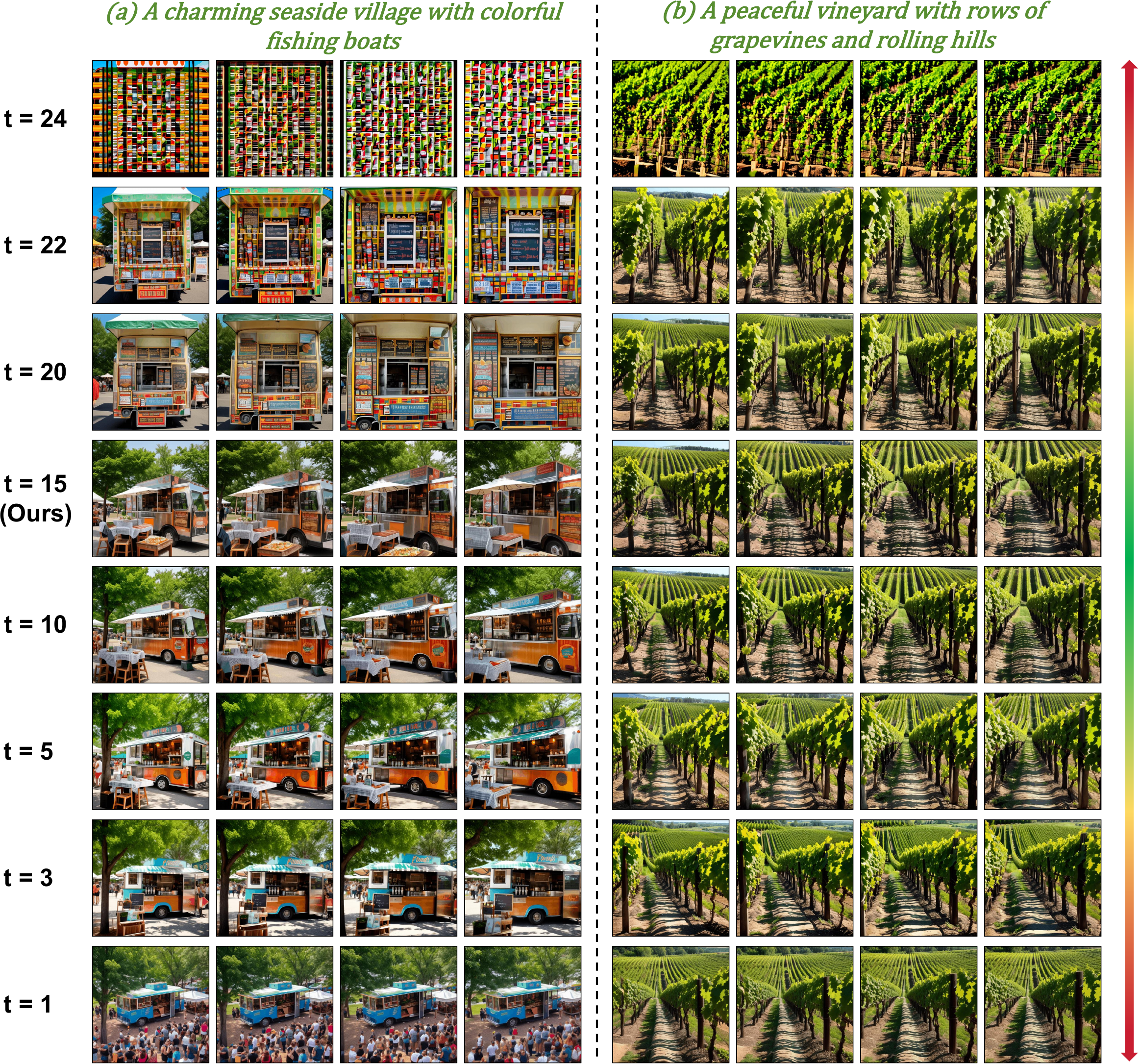}
\caption{Ablation on the hyperparameter timestep $t$. When it $t$ is too large ($t\ge 22$), there is too much noise in the temporal attention map, which causes the artifacts in the generated videos. When $t$ is too small ($t\le 3$), the latent $z_t$ is too close to the denoised $z_0$, where the temporal attention module contains less motion information. Therefore, the camera motions in the generated results are not as obvious as others. We choose the medium timestep $t=15$, whose temporal attention maps capture the video motions accurately.}
\label{fig:ablation on timestep}
\end{figure*}

In section 3.2 of the main paper, we propose that we can employ the temporal attention map in one intermediate step $t$ to represent the motions in different timesteps.  We find that the timestep $t$ cannot either be too large or too small, since the temporal attention map in a too large $t$ contains too much noise and the temporal attention map in a too small $t$ contains little temporal information. To validate this, we conduct one-shot camera motion transfer experiments on different timesteps $t$, which are shown in Fig.~\ref{fig:ablation on timestep}. It can be seen that when $t$ is too large ($t\ge 22$), the output videos suffer from heavy artifacts due to the noise in the temporal attention map. And when $t$ is too small ($t\le 3$), the generated video cannot be correctly generated since the temporal attention map contains too little motion information, which fails to guide the video generation process with accurate camera motions. The intermediate timesteps $5< t<20$ all generate good results. Therefore, we choose timestep $t=15$ as our default hyperparameter.

\section{User Study}
\label{sec:User Study}
In this section, we conduct a user study to evaluate the effectiveness of our MotionMaster. We have invited 28 volunteers in related research areas to rank the generated results from AnimateDiff+Lora~\cite{guo2023animatediff}, MotionCrtl~\cite{wang2023motionctrl} and our MotionMaster considering the generation quality, diversity, and the camera transfer accuracy. Specifically, each volunteer ranked 20 sets of results, where each basic camera motion (pan left, pan right, zoom in, and zoom out) contains 5 videos. We compute the average ranking and the percentage of ranking first of the three methods, which is shown in Tab.~\ref{tab:user study}. It can be seen that our model ranks first in \textbf{74.13\%} situations and achieves the best average rank of \textbf{1.27}, demonstrating the superior performance of our MotionMaster in camera motion transfer.

\begin{table}[t]
\setlength{\abovecaptionskip}{4pt}
\setlength{\belowcaptionskip}{-0.2cm}
\setlength\tabcolsep{3pt}
\renewcommand{\arraystretch}{1.2}
\caption{User Study from 28 volunteers.}
\resizebox{1.0\linewidth}{!}{
\begin{tabular}{c|ccc}
\toprule
Method & Animatediff+Lora & MotionCtrl & MotionMaster (Ours) \\
\midrule
\makecell[c]{Percentage of \\ Ranking First (\%) $\uparrow$} & 0.65\% & 25.22\% & \textbf{74.13\%}\\
Average Rank $\downarrow$ & 2.93 & 1.80 & \textbf{1.27} \\
\bottomrule
\end{tabular}}
\label{tab:user study}
\end{table}

